\documentclass{article} % For LaTeX2e
\usepackage{nips14submit_e,times}

\nipsfinalcopy % Uncomment for camera-ready version

\usepackage[ruled,vlined]{algorithm2e}
\usepackage{algorithmic}
\usepackage{amsfonts}
\usepackage{amsmath}
\usepackage{amssymb}
\usepackage{bm}
\usepackage{color}
\usepackage{graphicx}
\usepackage{hyperref}
\usepackage{mathrsfs}
\usepackage{textpos}
\usepackage{tikz}
\usepackage{url}
\usepackage{wrapfig}

\usepackage[font=small,labelfont=bf,tableposition=top]{caption}
\usepackage[font=footnotesize]{subcaption}

\usetikzlibrary{arrows,calc,shapes,snakes,automata,backgrounds,fit,petri}

\allowdisplaybreaks[1]

\def\const{\mathop{\rm const}}

\def\var{\mathop{\sf var}}

\def\pa{\mathrm{pa}}
\def\cp{\mathrm{cp}}
\def\ch{\mathrm{ch}}
\def\hidden{\mathrm{h}}
\def\observed{\mathrm{o}}
\def\drm{\mathrm{d}}

\def\Ebb{\mathbb{E}}

\def\bmeta{\bm{\eta}}

\def\blambda{\bm{\lambda}}
\def\bLambda{\mathbf{\Lambda}}

\def\0{\mathbf{0}}
\def\1{\mathbf{1}}

\def\c{\mathbf{c}}

\def\u{\mathbf{u}}
\def\v{\mathbf{v}}
\def\x{\mathbf{x}}

\def\U{\mathbf{U}}
\def\V{\mathbf{V}}

\def\X{\mathbf{X}}

\def\Ccal{\mathcal{C}}
\def\Dcal{\mathcal{D}}

\def\Lcal{\mathcal{L}}

\def\Ncal{\mathcal{N}}

\def\defined{\stackrel{.}{=}}%{\stackrel{\text{\tiny def}}{=}}
\newcommand{\appropto}{\mathrel{\vcenter{
  \offinterlineskip\halign{\hfil$##$\cr
    \propto\cr\noalign{\kern2pt}\sim\cr\noalign{\kern-2pt}}}}}

\definecolor{Blue}{rgb}{0.0,0.0,1.0}
\definecolor{Gray}{rgb}{0.5,0.5,0.5}

\title{On the Convergence of Stochastic Variational Inference in Bayesian Networks}

\author{
Ulrich Paquet \\
Microsoft Research \\
\texttt{ulripa@microsoft.com}}

\begin{document} 

\begin{textblock*}{150mm}(.0\textwidth,-2cm)
{\color{Gray}NIPS 2014 Workshop on
\emph{Advances in Variational Inference}. Montreal, Canada}
\end{textblock*}

\maketitle

\begin{abstract} 
We highlight a pitfall when applying stochastic variational inference to general Bayesian networks.
For global random variables approximated by an exponential family distribution,
natural gradient steps, commonly starting from a unit length step size, are averaged to convergence.
This useful insight into the scaling of initial step sizes is lost when the approximation factorizes across a general Bayesian network,
and care must be taken to ensure practical convergence.
We experimentally investigate how much of the baby (well-scaled steps) is thrown out with the bath water (exact gradients).
\end{abstract} 
\section{Introduction}
\label{sec:introduction}

Stochastic variational inference is framed as maximizing a global\footnote{The evidence lower bound is locally optimized with respect to local variational parameters.} variational parameter $\bLambda$, which is the natural parameter of a conjugate exponential distribution \cite{JMLR:hoffman13a}.
In this framework, stochastic gradient steps are taken along the natural gradient \cite{Amari_1998} to optimize for $\bLambda$.
A pleasing property of stochastic variational inference on a conjugate exponential distribution and approximation $q(\bLambda)$ is that 
the gradient is automatically rescaled so that a unit-length step size will minimize it.
For a general Bayesian network, where the global variational parameters are subdivided to parameterize different factors $q_i$ in the network's variational approximation, the picture is less clear.
Hoffman \emph{et al.}'s appendix suggests a stochastic updating scheme like that of the global version \cite{JMLR:hoffman13a}.
We show here that the problem is more subtle in the general case, as component-wise noisy natural gradients
can tightly couple variational parameters,
and following the default recipe can sometimes lead to a scheme that ``diverges'' beyond recovery!

These remarks are of particular value to the Xbox recommender system,
which uses stochastic variational inference in a Bayesian network on ``worldwide'' scale \cite{xbox-www, PaquetK13}.
Some of the results presented in Sec.~\ref{sec:bmf} are preliminary investigations
that were done when designing the system in 2012.

\section{Variational Bayes}

A Bayesian network between the variables $\X = \{ \x_j \}$ defines the conditional dependency structure between them through their joint probability
$
p(\X) = \prod_j p(\x_j | \pa_j)
$.
Following Fig.~\ref{fig:bayesnet},
let $\pa(j)$ be the set of indexes of parents of random variable(s) $\x_j$;
for notational convenience we let $\pa_j \defined \{ \x_k : k \in \pa(j) \}$ denote the parent variables.
The variables in the network can be hidden or observed, $\X = \{ \X^\hidden, \X^\observed \}$.
Variational Bayes (VB) approximates the posterior $p(\X^\hidden | \X^\observed)$ with $q(\X^\hidden)$,
by maximizing the evidence lower bound
\[
\Lcal[q] \defined \int  q(\X^\hidden) \log \frac{p(\X)}{q(\X^\hidden)} \, \drm \X^\hidden 
\le \log p(\X^\observed) \ .
\]
If $i$ indexes the hidden variables, we factorize the approximation with
\[
q(\X^\hidden) = \prod_i q_i(\x_i) \ .
\]
Let $\Ebb_{j \neq i}$
indicate the expectation taken over $\prod_{j \neq i} q_j(\x_j)$.
The bound can be maximized in a component-wise fashion by iteratively setting
each $q_i(\x_i)$ to the maximum
\begin{equation} \label{eq:vbupdate}
\log q_i^*(\x_i)  = \Ebb_{j \neq i} \Big[ \log p(\x_i | \pa_i) \Big]
+ \sum_{k \in \ch(i)} \Ebb_{j \neq i} \Big[ \log p(\x_k | \pa_k) \Big] + \const \ .
\end{equation}
In many practical networks there are some $\x_i$ for which number of children $N_i \defined |\ch(i)|$ is large.
In \cite{JMLR:hoffman13a}, $\x_i$ is a topic-vocabulary distribution from which millions of documents are generated.
In Sec.~\ref{sec:bmf} and \cite{xbox-www, PaquetK13} the interaction is bilinear,
where user $\x_i$ and item $\x_j$ variables are combined to represent a user's affinity to an item.
Rather than summing over all $\ch(i)$ for each update in (\ref{eq:vbupdate}), we aim to stochastically approximate the expectations. 
It alleviates two problems: firstly, the sum contains many terms; secondly, the update depends on some $q(\x_j)$ which will be re-estimated, and the expense of fully estimating $q(\x_i)$ is lost as it too will be re-estimated.

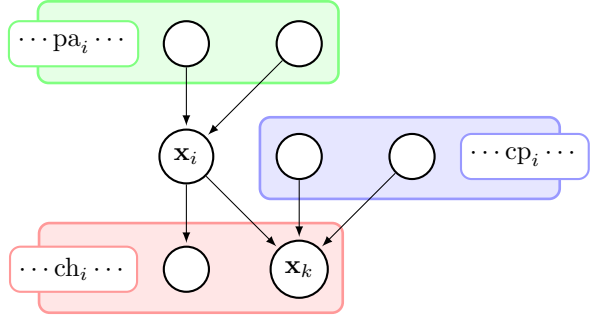
\begin{figure}[t]
\begin{minipage}[b]{0.6\linewidth}
\begin{center}
\begin{tikzpicture}[bend angle=45,>=latex]
  \tikzstyle{obs} = [ circle, thick, draw = black!100, fill = blue!20, minimum size = 6mm ]
  \tikzstyle{lat} = [ circle, thick, draw=black!100, fill = red!0, minimum size = 6mm ]
  \tikzstyle{par} = [ circle, draw, fill = black!100, minimum width = 3pt, inner sep = 0pt]
  \tikzstyle{parents} = [ rectangle, thick, draw=green!50, fill = green!0, minimum size = 6mm ]
  \tikzstyle{children} = [ rectangle, thick, draw=red!40, fill = red!0, minimum size = 6mm]
  \tikzstyle{copar} = [ rectangle, thick, draw=blue!40, fill = blue!0, minimum size = 6mm]	
  \tikzstyle{every label} = [black!100]
  %\tikzstyle{every edge} = [black!90]
  \begin{scope}[node distance = 1.5cm and 1.5cm,rounded corners=4pt]
    \node [lat] (xi) {$\x_i$};
    \node [lat] (xpar1) [above of = xi] { }
      edge [post] (xi);
    \node [lat] (xpar2) [right of = xpar1] { }
      edge [post] (xi);
    \node [parents] (xpar3) [left of = xpar1] { $ \cdots \pa_i \cdots $ };
    \node [lat] (xch1) [below of = xi] { }
      edge [pre] (xi);
    \node [lat] (xch2) [right of = xch1] {$\x_k$ }
      edge [pre] (xi);
    \node [children] (xch3) [left of = xch1] { $\cdots \ch_i \cdots $ };    
    \node [lat] (xcp1) [right of = xi] { }
      edge [post] (xch2);
    \node [lat] (xcp2) [right of = xcp1] { }
      edge [post] (xch2);
    \node [copar] (copar) [right of = xcp2] { $\cdots \cp_i \cdots$ };
        
    \begin{pgfonlayer}{background}
    \filldraw [line width = 1pt, draw=green!50, fill=green!10]
	($(xpar3.north west) + (0.4cm, 0.25cm)$) rectangle ($(xpar2.south east) + (0.3cm, -0.3cm)$);
     \filldraw [line width = 1pt, draw=red!40, fill=red!10]
	 ($(xch3.north west) + (0.4cm, 0.3cm)$) rectangle ($(xch2.south east) + (0.3cm, -0.3cm)$);
	 \filldraw [line width = 1pt, draw=blue!40, fill=blue!10]
	 ($(xcp1.north west) + (-0.3cm, 0.3cm)$) rectangle ($(copar.south east) + (-0.4cm, -0.25cm)$);	 
     \end{pgfonlayer}
   \end{scope}
\end{tikzpicture}	
\end{center}
\end{minipage}
\begin{minipage}[b]{0.39\linewidth}
\caption{A Bayesian network, indicating $\x_i$'s Markov blanket. The parents of $\x_i$ are $\pa_i$ and its
children $\x_k \in \ch_i$. For a compact notation we also write $k \in \ch(i)$ as the index set of children, where it is clear from context. Each child $k$ has parents $\x_i$ and $\cp_i$ (the co-parents with $\x_i$).
The form of our notation loosely matches Winn and Bishop's in \cite{Winn_Bishop_2005}, as Alg.~\ref{alg:stochastic}
can be interpreted as \emph{``stochastic variational message passing''}.
}
\label{fig:bayesnet}
% \vspace{5pt}
\end{minipage}
\vspace{-5pt}
\end{figure}

% \vspace{-3pt}
\subsection{Conditionally conjugate models}

The updates in (\ref{eq:vbupdate}) are straightforward when the Bayesian network is conditionally conjugate; that is, when the distribution of $\x_i$, conditioned on $\pa_i$, is (a) drawn from an exponential family, and (b) is conjugate with respect to the distribution of $\pa_i$.
We define the exponential family as
\begin{equation} \label{eq:exponential}
\log p(\x_i | \pa_i) = \bmeta_i(\pa_i)^T \phi_i(\x_i) + f_i(\x_i) + g_i(\pa_i)
\end{equation}
where $\bmeta_i(\pa_i)$ is the natural parameter vector, $\phi_i(\x_i)$ forms the sufficient
statistics, and $g_i(\pa_i)$ defines the normalizing constant through
$
g_i(\pa_i) = - \log \int  \exp \{ \bmeta_i(\pa_i)^T \phi_i(\x_i) + f_i(\x_i) \} \, \drm \x_i
$.
We can view (\ref{eq:exponential}) as a ``prior'' over $\x_i$.
Now consider a node $\x_k \in \ch_i$ in Fig.~\ref{fig:bayesnet}. We subdivide $\pa_k$, the parents of $\x_k$, into $\x_i$ and its co-parents $\cp_i$:
\[
\log p(\x_k | \x_i, \cp_i) = \bmeta_{k}(\x_i, \cp_i)^T \phi_k(\x_k) + f(\x_k) + g(\x_i, \cp_i) \ .
\]
We can view this as a contribution to the ``likelihood'' of $\x_i$. We include the co-parents as they are part of $\x_i$'s Markov blanket.
Through conjugacy, $p(\x_i | \pa_i)$ and $p(\x_k | \x_i, \cp_i)$ have the same functional form with respect to $\x_i$, so that we can
rewrite $p(\x_k | \x_i, \cp_i)$ in terms of the sufficient statistics $\phi_i(\x_i)$ by defining some function $\bmeta_{ki}$ with
\[
\log p(\x_k | \x_i, \cp_i) = \bmeta_{ki}(\x_k, \cp_i)^T \phi_i(\x_i) + h(\x_k, \cp_i) \ .
\]
We furthermore parameterize the $q(\x_i)$ distributions in terms of their natural parameters. To distinguish them, we denote their natural parameters by $\blambda_i$, and define $\bLambda \defined [\blambda_1, \ldots, \blambda_I ]$:
\begin{equation} \label{eq:q}
\log q_i(\x_i | \blambda_i) = \blambda_i^T \phi_i(\x_i) + f_i(\x_i) + \tilde{g}_i(\blambda_i) \ .
\end{equation}

% \vspace{-3pt}
\subsection{Variational Bayes updates and their stochastic version}

Returning to (\ref{eq:vbupdate}), we can write
\begin{equation}
\log q_i^*(\x_i) = \Ebb_{j \neq i} \Bigg[  \bmeta_i(\pa_i) 
+ \sum_{k \in \ch(i)} \bmeta_{ki}(\x_k, \cp_i)  \Bigg]^T \phi_i(\x_i)
+ f_i(\x_i) + \const \ , \label{eq:vbconjugate}
\end{equation}
from which we can directly read off the updated natural parameter $\blambda_i^*$ through (\ref{eq:q}).
Notice now that $\bmeta_i$ is a multi-linear function of the random variables $\pa_i$, i.e.~it is linear in \emph{each} parent random variable.
In the same way $\bmeta_{ki}$ is a multi-linear function of the random variables $\x_k$ and $\cp_i$.  Furthermore, $q$ factorizes over these variables (except where they are observed, of course). We can therefore reparameterize (\ref{eq:vbconjugate}) in terms of expectations over $q_j$, $i \neq j$ with
\begin{align*}
\Ebb_{j \neq i} \Big[  \bmeta_i(\pa_i) \Big] & =
\widetilde{\bmeta}_i \left( \Big\{ \Ebb_{j} \Big[ \phi_j(\x_j) \Big] \Big\}_{j \in \pa(i)} \right)
\defined \widetilde{\bmeta}_i \\
\Ebb_{j \neq i} \Big[  \bmeta_{ki}(\x_k, \cp_i) \Big] & = 
\widetilde{\bmeta}_{ki} \left( \Ebb_{k} \Big[ \phi_k(\x_k) \Big], \,  \Big\{ \Ebb_{j} \Big[ \phi_j(\x_j) \Big] \Big\}_{j \in \cp(i)} \right)
\defined \widetilde{\bmeta}_{ki} \ .
\end{align*}

\begin{wrapfigure}[12]{L}[0pt]{0pt}
\noindent\begin{minipage}[t]{0.51\textwidth}
\vspace{-14pt}
\begin{algorithm}[H]
   \caption{Stochastic Variational Bayes}
   \label{alg:stochastic}
\begin{algorithmic}[1]
   \FOR{$t=1$ to $t_{\max}$ or convergence}
   \STATE $\rho_t = (t + \tau)^{-\kappa}$
   \FOR{each hidden $\x_i$}
   \STATE $\Ccal \leftarrow C$ random nodes from $\ch_i$
   \STATE $\blambda_i^{\mathrm{temp}} \leftarrow \widetilde{\bmeta}_i + \frac{N_i}{C} \sum_{k \in \Ccal}  \widetilde{\bmeta}_{ki}$
   \STATE \emph{option (a):} $\blambda_i \leftarrow (1 - \rho_t ) \blambda_i + \rho_t \blambda_i^{\mathrm{temp}}$
   \ENDFOR
   \STATE \emph{option (b):} $\bLambda \leftarrow (1 - \rho_t ) \bLambda + \rho_t \bLambda^{\mathrm{temp}}$
   \ENDFOR
\end{algorithmic}
\end{algorithm}
\end{minipage}
\end{wrapfigure}

This is a key ingredient of algorithms like variational message passing \cite{Winn_Bishop_2005}.
(When $\x_k$ is observed, $\phi_k(\x_k)$ is kept as is, as it is averaged over a delta function.) 
The variational update in (\ref{eq:vbconjugate}) becomes
\begin{equation} \label{eq:vb-update}
\blambda_i^{*} = \widetilde{\bmeta}_i + \sum_{k \in \ch(i)} \widetilde{\bmeta}_{ki} \ .
\end{equation}
The update is a step along the natural gradient \cite{Amari_1998}, equivalent to setting the gradient to zero by solving for the zero of the derivative of $\Lcal$ with respect to $\blambda_i$.
In particular, (\ref{eq:vb-update}) updates $\blambda_i$ from its old value to $\blambda_i^{*}$ using a \emph{step of unit length}
along the natural gradient.
Sec.~\ref{sec:component-grad} derives the gradient $\nabla_{\blambda_i} \Lcal$,
and Sec.~\ref{sec:component-natural-grad} states its natural form $\widehat{\nabla}_{\blambda_i} \Lcal$.

When $N_i \defined |\ch(i)|$ is large, not all the child nodes might be accessed in reasonable time.
Furthermore, when $q(\x_i)$ is re-estimated, the (previous) large computation is discarded and recomputed.
We may alternatively consider a subsample of nodes from $\ch(i)$ to determine the sufficient statistics.
By placing a uniform distribution $\widetilde{p}_i$ on the atoms $\widetilde{\bmeta}_{ki}$, the update from (\ref{eq:vb-update}) is equivalent to
$
\blambda_i = \widetilde{\bmeta}_i + N_i \, \Ebb_{\widetilde{p}_i} [ \widetilde{\bmeta} ]
$.
This expectation can be estimated in many ways.
Let set  $\Ccal$ be a sample of $C$ children from $\ch_i$ without replacement and let
\begin{equation} \label{eq:stochasticupdate}
\blambda_i^{\mathrm{temp}} = \widetilde{\bmeta}_i + \frac{N_i}{C} \sum_{k \in \Ccal}  \widetilde{\bmeta}_{ki} \ .
\end{equation}
Taking expectations gives $\blambda_i^{*} = \Ebb[ \blambda_i^{\mathrm{temp}} ] = \widetilde{\bmeta}_i + N_i  \Ebb_{\widetilde{p}_i} [ \widetilde{\bmeta} ]$.
With $\rho_t \to 0$, $\sum_{t=1}^{\infty} \rho_t = \infty$ and $\sum_{t=1}^{\infty} \rho_t^2 < \infty$, these stochastic
natural gradients are used in Alg.~\ref{alg:stochastic}, which is a stochastic version of variational message passing.
In Alg.~\ref{alg:stochastic}, scalar $\kappa \in (\frac{1}{2}, 1]$ is a forgetting rate, while delay $\tau \ge 0$ discounts early iterations more.

There are two options in Alg.~\ref{alg:stochastic}. For option \emph{(a)},
the mean value of the parameters of $q(\X^{\hidden})$ is periodic in $I$, the number of factors in $q$, and convergence to a local optimum can also be guaranteed for $I$-dependent mean values \cite{Kushner_2003}. Option \emph{(b)} is the update scheme given in \cite{JMLR:hoffman13a}.

%%%%%%%%%%%%%%%%%%%%%%%%%%%%%%%%%%%%%%%%%%%%%%%%%%%%%%%
\section{Bayesian matrix factorization} \label{sec:bmf}

To illustrate a general Bayesian network,
we factorize a sparse matrix of a subsample of a million entries in the Netflix data set ($M = 4805$ users and $N = 16015$ items). Each entry $r_{mn}$ is user $m$'s rating of movie $n$ on a five-star rating scale. For illustrative purposes, consider a Gaussian bilinear ratings model
\[
p(r_{mn} | \u_m, \v_n) = \Ncal( r_{mn} \, ; \, \u_m^T \v_n , \, 1)
\]
for user parameter vector $\u_m \in \mathbb{R}^K$ and item trait vector $\v_n \in \mathbb{R}^K$.
We place a factorized prior $\Ncal(u_{mk} \, ; \, 0, 1)$ on each of the entries of $\u_m$ and $\v_n$.
We choose a fully factorized Gaussian approximation $q(\U) = \prod_m \prod_k q(u_{mk})$, with a similar approximation for $q(\V)$.
The VB update for $q(u_{mk})$ therefore incorporates $2K-1$ co-parents due to the inner product.
With the Gaussian's natural parameters being its precision and mean-times-precision, it is
\begin{equation} \label{eq:bmf}
\blambda_{mk}^{\mathrm{temp}} =
\begin{pmatrix}
\mathsf{prec} \\
\mathsf{mean} \cdot \mathsf{prec}
\end{pmatrix}
=
\begin{pmatrix} 1 \\ 0 \end{pmatrix}
+ \frac{N_{m}}{C}\sum_{n \in \Ccal}
\begin{pmatrix}
\var_q[v_{nk}] + \Ebb_q[v_{nk}]^2 \\
\Ebb_q[v_{nk}] (r_{mn} - \sum_{k' \neq k} \Ebb_q[u_{mk'}] \Ebb_q[v_{nk'}] )
\end{pmatrix} \ .
\end{equation}

\begin{figure}[t]
%\vskip 0.2in
\begin{center}
\includegraphics[width=0.49\textwidth]{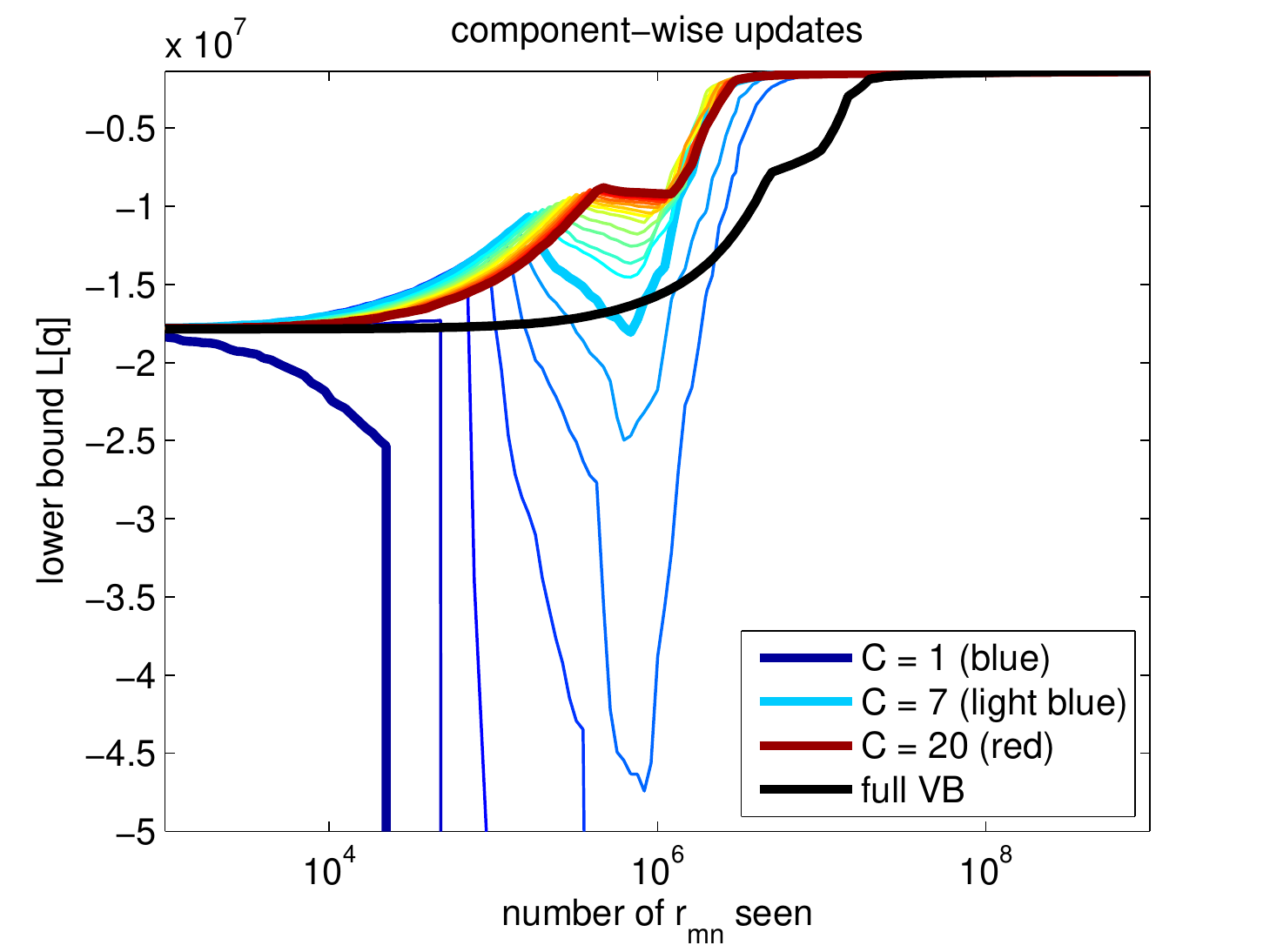}
\includegraphics[width=0.49\textwidth]{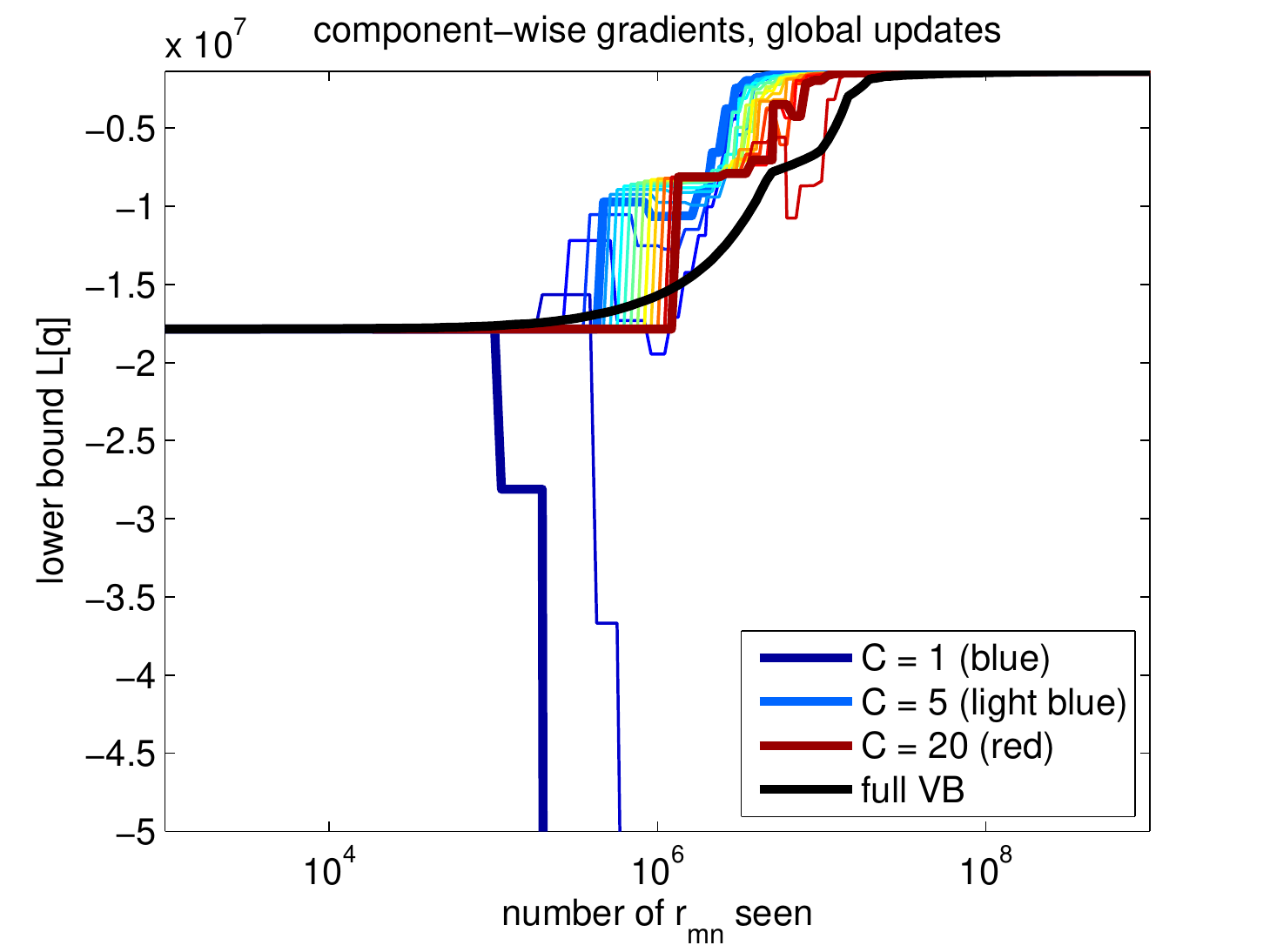} \\
\includegraphics[width=0.49\textwidth]{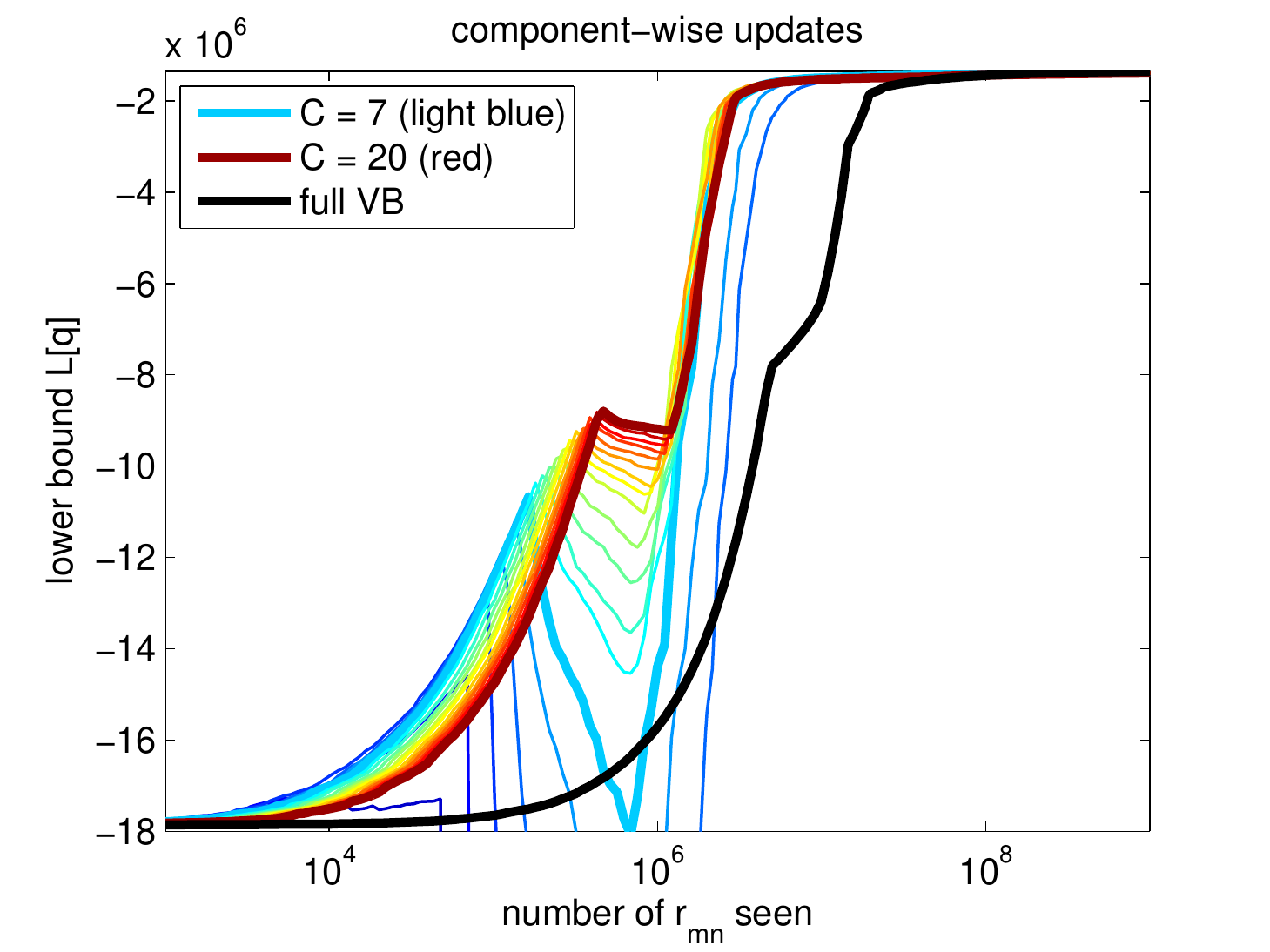}
\includegraphics[width=0.49\textwidth]{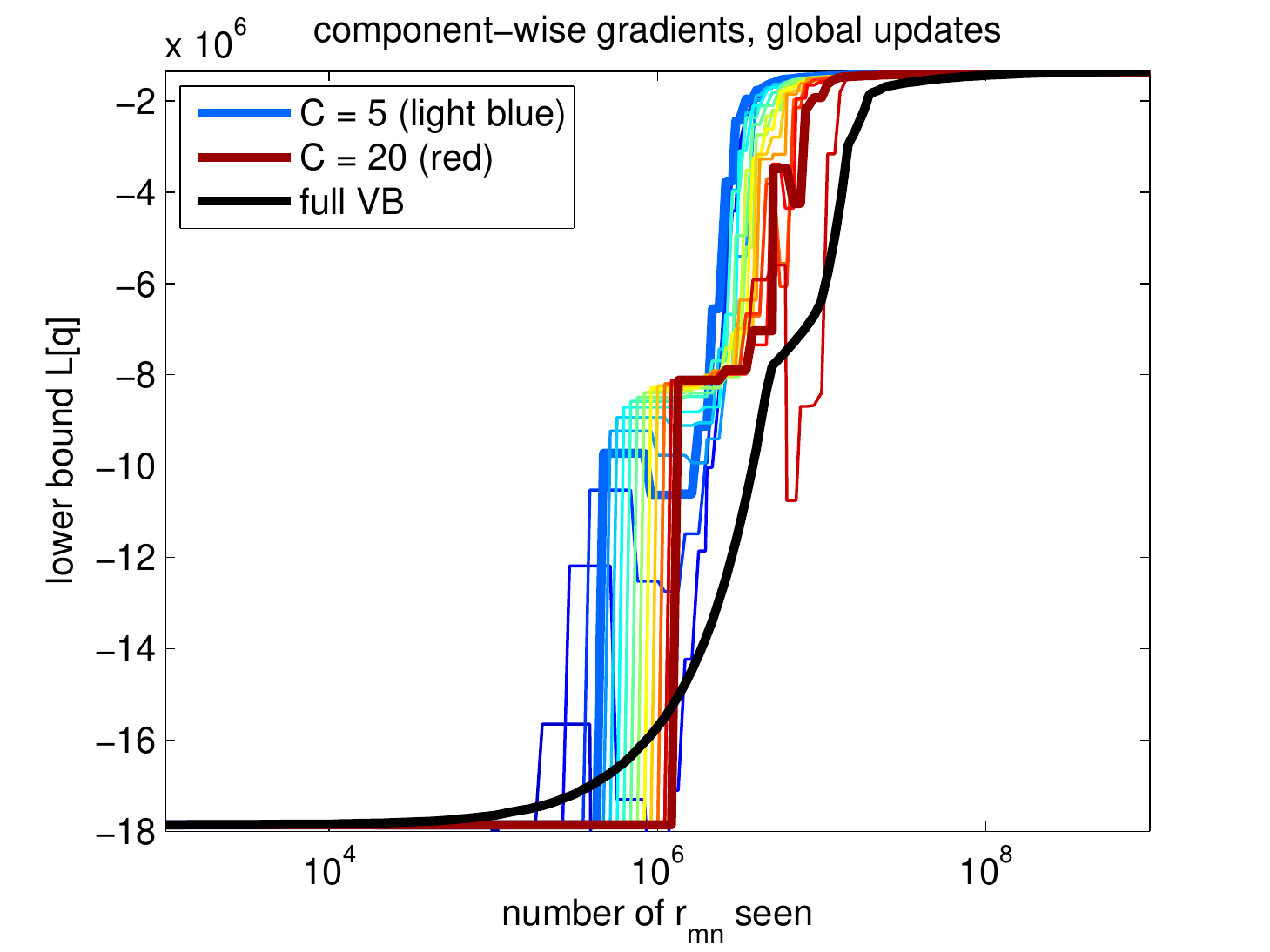} \\
\includegraphics[width=0.49\textwidth]{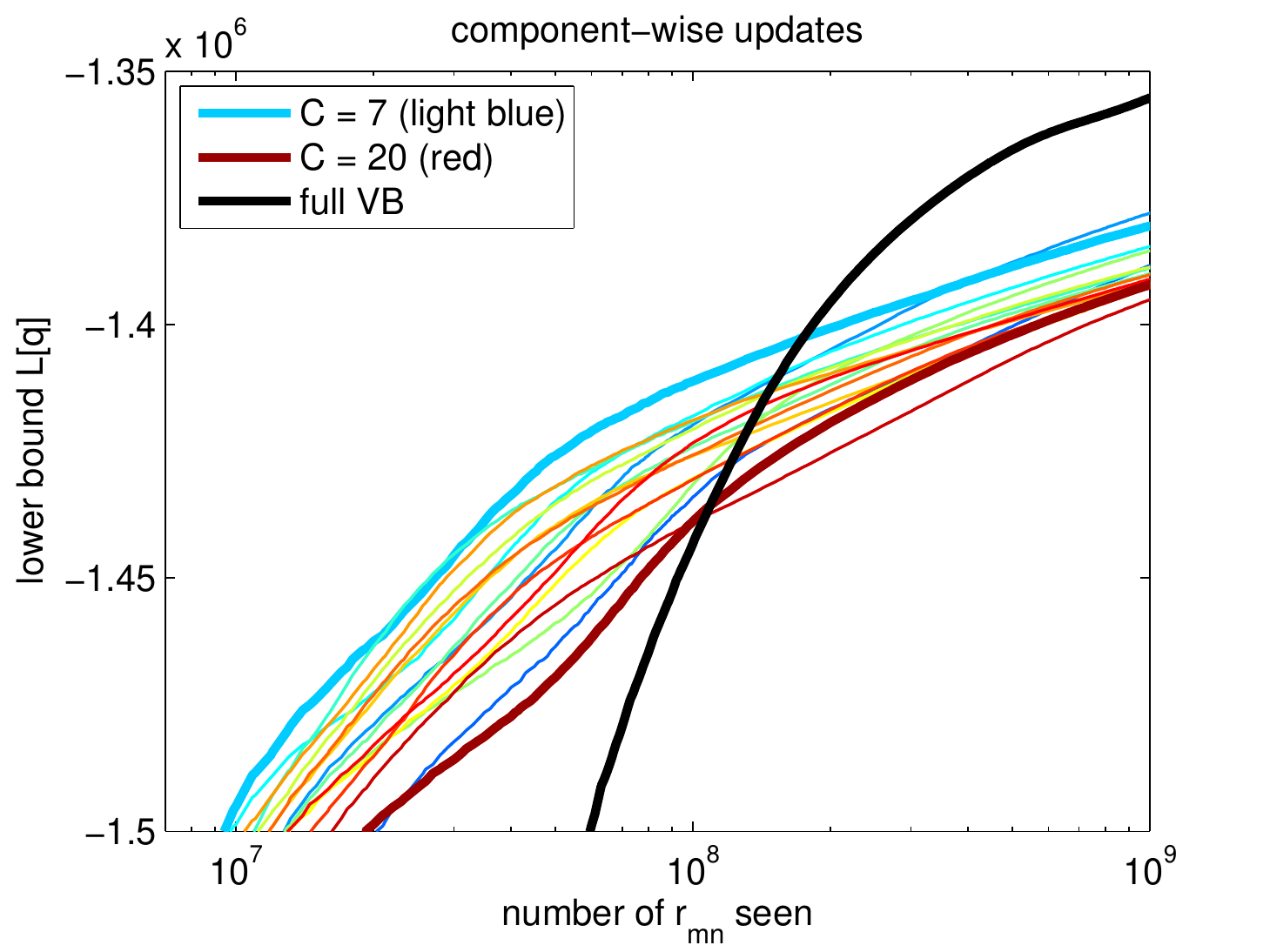}
\includegraphics[width=0.49\textwidth]{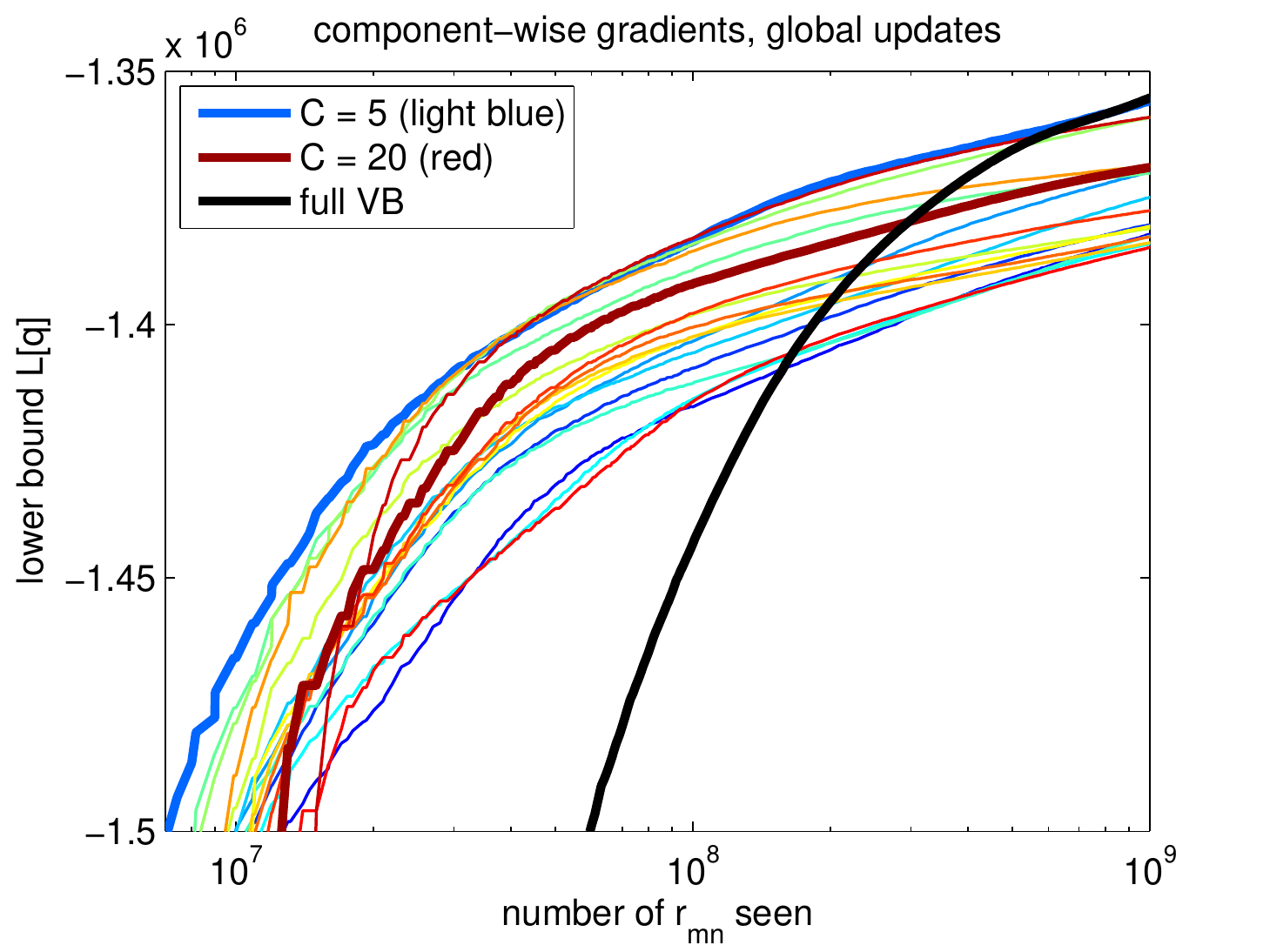}
\end{center}
\caption{Convergence of ${\cal L}[q]$ with $\rho_t = t^{-0.6}$. Alg.~\ref{alg:stochastic}'s option \emph{(a)} is shown in the left
column; option \emph{(b)} is shown in the right column. The x-axes are on a logarithmic scale.
The \emph{global} stochastic gradient is not in its natural form,
and the effect of a large variance in the gradient estimate
and overshooting with too large step sizes of $\rho_t \in (0, 1]$ is clearly visible for small $C$.
Note that $r_{mn}$'s can be revisited over multiple loops in Alg.~\ref{alg:stochastic}.
Different magnifications of the same two convergence plots for options \emph{(a)} and \emph{(b)} are shown in the three rows of graphs.
}
\label{fig:convergence}
\vspace{-18pt}
\end{figure} 

Fig.~\ref{fig:convergence} shows ${\cal L}[q]$ as a function of the number of times that individual ratings (observed nodes) $r_{mn}$ are accessed or queried (using $K=5$).
The value of the bound is shown for the use of at most $C = 1, \ldots, 20$ children when estimating the gradient of each random variable with (\ref{eq:stochasticupdate}) and (\ref{eq:bmf}). Both options \emph{(a)} and \emph{(b)} in Alg.~\ref{alg:stochastic} ``diverge'' in a numerically unrecoverable way when $C$ is small.
This is due to the global gradient not being in its natural form, and using a step size of $\rho_t \in (0, 1]$ that is too big, overshooting with too large gradient steps.

Full VB, shown in black in Fig.~\ref{fig:convergence}, implicitly uses $\rho_t = 1$ in (\ref{eq:vb-update}).
As the stochastic natural gradient depends on other $\blambda_j$, much smaller initial step sizes are required to not ``overshoot''.
The variance of the gradient is simply too large compared to $\rho_t$.
Fig.~\ref{fig:convergence} illustrates this problem for \emph{general Bayesian networks};
see especially the \emph{top left} figure.

In practice, we can overcome this problem overcome by starting with sufficiently small initial step sizes $\rho_t \ll 1$.
For $C = 1$ in option \emph{(a)} this was starting from $\rho_1 = \frac{1}{512}$,
and $\rho_1 = \frac{1}{64}$ for option \emph{(b)}. In \cite{xbox-www, PaquetK13} the value of $C$ varied depending on a user or item's usage,
and there $\rho_t = 1$ was fixed for the first ten iterations before slowly decreasing it.

Have we thrown the baby (well-scaled steps) out with the bath water (exact gradients)?
Maybe some. As shown by this short note, it is still an open question.

% Use unnumbered third level heading for the references.
\renewcommand\section{\subsubsection}

%%%%%%%%%%%%%%%%%%%%%%%%%
\section*{Acknowledgment}

To the anonymous reviewer who pointed out that the Fisher information matrix of $q(\X^\hidden | \bLambda)$ block-diagonal, having
the Fisher information matrices of $q(\x_i | \blambda_i)$ along its diagonal: thank you!

% Any choice of citation style is acceptable as long as you are consistent. It is permissible to reduce
% the font size to ‘small’ (9-point) when listing the references. Remember that this year you can use
% a ninth page as long as it contains only cited references.
\small
% \bibliographystyle{plain}
% \bibliography{nips}

\begin{thebibliography}{1}

\bibitem{Amari_1998}
S.~Amari.
\newblock Natural gradient works efficiently in learning.
\newblock {\em Neural Computation}, 10(2):251--276, 1998.

\bibitem{JMLR:hoffman13a}
M.~D. Hoffman, D.~M. Blei, C.~Wang, and J.~Paisley.
\newblock Stochastic variational inference.
\newblock {\em Journal of Machine Learning Research}, 14:1303--1347, 2013.

\bibitem{Kushner_2003}
H.~J. Kushner and G.~G. Yin.
\newblock {\em Stochastic Approximation and Recursive Algorithms and
  Applications}.
\newblock Springer, 2003.

\bibitem{xbox-www}
U.~Paquet and N.~Koenigstein.
\newblock One-class collaborative filtering with random graphs.
\newblock In {\em Proceedings of the 22nd international conference on World
  Wide Web}, WWW '13, pages 999--1008, 2013.

\bibitem{PaquetK13}
U.~Paquet and N.~Koenigstein.
\newblock One-class collaborative filtering with random graphs: Annotated
  version.
\newblock {\em arXiv:1309.6786}, 2013.

\bibitem{Winn_Bishop_2005}
J.~Winn and C.M. Bishop.
\newblock Variational message passing.
\newblock {\em Journal of Machine Learning Research}, 6:661--694, 2005.

\end{thebibliography}

\normalsize

% Get the main heading back to how it was.
\makeatletter
\renewcommand\section{\@startsection {section}{1}{\z@}{-2.0ex plus -0.5ex minus -.2ex}{1.5ex plus 0.3ex minus0.2ex}{\large\bf\raggedright}}
\makeatother

\appendix

%%%%%%%%%%%%%%%%%%%
\section{Gradients}

In this Appendix, we derive the component-wise gradients and their natural version, and present basic intuition for why steps down the stochastic gradient can be taken.

\subsection{Component-wise gradients} \label{sec:component-grad}

The function that's minimized to find (\ref{eq:vb-update}) is $\Lcal(\blambda_i)$ below. It is a function of $\blambda_i$, whilst keeping all other $\blambda_{j}$ for $j \neq i$ fixed:
\begin{align*}
\Lcal(\blambda_i) & = \Ebb_q \Bigg[ \log p(\x_i | \pa_i) + \sum_{k \in \ch(i)}\log p(\x_k | \x_i, \cp_i) 
- \log q_i(\x_i | \blambda_i) \Bigg]
\\
& =
\Ebb_q \Bigg[ \bmeta_i(\pa_i)^T \phi_i(\x_i) + f_i(\x_i) + g_i(\pa_i) \\
& \qquad\qquad + \sum_{k \in \ch(i)} \Big( \bmeta_{k}(\x_i, \cp_i)^T \phi_k(\x_k) + f(\x_k) + g(\x_i, \cp_i) \Big) \\
& \qquad\qquad - \blambda_i^T \phi_i(\x_i) - f_i(\x_i) - \tilde{g}_i(\blambda_i) \Bigg]
\end{align*}
Because of local conjugacy, $p(\x_k | \x_i, \cp_i)$ can be rewritten in terms of the sufficient statistics $\phi_i(\x_i)$
through a multi-linear function $\bmeta_{ki}$ of the random variables $\x_k$ and $\cp_i$ to yield
\begin{align*}
\Lcal(\blambda_i)
& =
\Ebb_q \Bigg[ \bmeta_i(\pa_i)^T \phi_i(\x_i) + f_i(\x_i) + g_i(\pa_i) 
 + \sum_{k \in \ch(i)} \Big( \bmeta_{ki}(\x_k, \cp_i)^T \phi_i(\x_i) + h(\x_k, \cp_i) \Big) \\
& \qquad\qquad - \blambda_i^T \phi_i(\x_i) - f_i(\x_i) - \tilde{g}_i(\blambda_i) \Bigg] \ .
\end{align*}
Taking expectations over $q$ gives, as function of $\blambda_i$,
\begin{equation} \label{eq:L-of-lambda-i}
\Lcal(\blambda_i) =  \Bigg( \widetilde{\bmeta}_i + \sum_{k \in \ch(i)} \widetilde{\bmeta}_{ki} \Bigg)^T \Ebb_i \left[ \phi_i(\x_i) \right]
- \blambda_i^T \Ebb_i \left[ \phi_i(\x_i) \right] - \tilde{g}_i(\blambda_i)
+ \mathrm{const} \ ,
\end{equation}
with
$
\tilde{g}_i(\blambda_i) = - \log \int \exp \{ \blambda_i^T \phi_i(\x_i) + f_i(\x_i) \} \, \drm \x_i
$.
The derivatives of the log partition function $- \tilde{g}_i(\blambda_i)$ with respect to $\blambda_i$ give the expected sufficient statistics
\begin{align*}
- \nabla \tilde{g}_i(\blambda_i)
& = \Ebb_i \left[ \phi_i(\x_i) \right] \\
- \nabla^2 \tilde{g}_i(\blambda_i) 
& = \nabla \Ebb_i \left[ \phi_i(\x_i) \right] \\
& = \Ebb_i \left[ \Big( \phi_i(\x_i) - \Ebb_i \left[ \phi_i(\x_i) \right]  \Big)
\Big( \phi_i(\x_i) - \Ebb_i \left[ \phi_i(\x_i) \right]  \Big)^T \right] \\
& = {\sf cov}_i \left[ \phi_i(\x_i) \right] \ ,
\end{align*}
and by using properties of the exponential family, the gradient of $\Lcal$ with respect to $\blambda_i$ is therefore
\begin{equation} \label{eq:grad}
\nabla_{\blambda_i} \Lcal(\blambda_i) = {\sf cov}_i \left[ \phi_i(\x_i) \right] \Bigg( \widetilde{\bmeta}_i + \sum_{k \in \ch(i)} \widetilde{\bmeta}_{ki} - \blambda_i \Bigg) \ .
\end{equation}
Solving for $\nabla_{\blambda_i} \Lcal(\blambda_i) = \0$ yields the component-wise
VB update $\blambda_i^{*} = \widetilde{\bmeta}_i + \sum_{k \in \ch(i)} \widetilde{\bmeta}_{ki}$ 
that we find in (\ref{eq:vb-update}).
Gradient $\nabla_{\blambda_i} \Lcal$ depends on $\blambda_i$ through ${\sf cov}_i [ \phi_i(\x_i) ]$ and $\blambda_i$,
and in the next section we will show that the natural gradient $\widetilde{\nabla}_{\blambda_i} \Lcal$ removes the
dependency on ${\sf cov}_i [ \phi_i(\x_i) ]$, so that it is a linear function of $\blambda_i$,
with the minimum being attained by taking a step of length one along it.

%%%%%%%%%%%%%%%%%%%%%%%%%%%%%%%%%%%%%%%%%%%%%%%%%%%%%%%%%%%%%%%%%%%%%%%%%%%%%%%%
\subsection{Component-wise natural gradients} \label{sec:component-natural-grad}

The Fisher information matrix of $q_i$ is
\begin{align*}
G(\blambda_i)
& = \Ebb_i \left[ \Big(\nabla_{\blambda_i} \log q (\x_i | \blambda_i) \Big) \Big( \nabla_{\blambda_i} \log q (\x_i | \blambda_i) \Big)^T \right] \\
& = \Ebb_i \left[ \Big( \phi_i(\x_i) - \Ebb_i[\phi_i(\x_i)] \Big) \Big( \phi_i(\x_i) - \Ebb_i[\phi_i(\x_i)] \Big)^T \right] \\
& = {\sf cov}_i \left[ \phi_i(\x_i) \right] \ ,
\end{align*}
and the component-wise natural gradient is obtained
by multiplying it with $\nabla_{\blambda_i} \Lcal$, yielding
\begin{equation} \label{eq:natgrad}
\widehat{\nabla}_{\blambda_i} \Lcal(\blambda_i)
\defined G(\blambda_i)^{-1} \nabla_{\blambda_i} \Lcal(\blambda_i) \nonumber \\
= \widetilde{\bmeta}_i + \sum_{k \in \ch(i)} \widetilde{\bmeta}_{ki} - \blambda_i \ . 
\end{equation}
A gradient descent along the natural gradient is taken with
step length $\rho > 0$. Starting at point $\blambda_i^{(t-1)}$, gradient descent updates it to $\blambda_i^{(t)}$ with
\begin{align*}
\blambda_i^{(t)}
\leftarrow \blambda_i^{(t-1)} + \rho \widehat{\nabla}_{\blambda_i^{(t-1)}} \Lcal(\blambda_i)
& = \blambda_i^{(t-1)} + \rho \Bigg( \widetilde{\bmeta}_i + \sum_{k \in \ch(i)} \widetilde{\bmeta}_{ki} - \blambda_i^{(t-1)} \Bigg) \\
& = (1 - \rho) \blambda_i^{(t-1)} + \rho \Bigg( \widetilde{\bmeta}_i + \sum_{k \in \ch(i)} \widetilde{\bmeta}_{ki} \Bigg) \\
& = (1 - \rho) \blambda_i^{(t-1)} + \rho \blambda_i^{*} \ .
\end{align*}
When the above update is compared to (\ref{eq:vb-update}), we see that
the minimum $\blambda_i^{(t)} \leftarrow \blambda_i^{*}$ is obtained by applying a step size of $\rho = 1$ along the natural gradient.

%%%%%%%%%%%%%%%%%%%%%%%%%%%%%%%%%%%%%%%%%%%%%%%%%%%%%%%%%%%%%%%%%%%%%%%%%%%%%%%%%%%%%%%%
\subsection{Stochastic natural gradients: a bird's eye view} \label{sec:stochastic-grad}

In this section an intuitive motivation will be provided for doing stochastic gradient descent using the natural gradient, as it was defined above in Sec.~\ref{sec:component-natural-grad}. The explanation favours an intuitive understanding above mathematical rigour.
Imagine that instead of $\blambda_i^{*}$, we have access to a sequence of samples 
$\{ \blambda_{i}^{\mathrm{temp},\tau} \}_{\tau = 1}^{t}$, so that
$\Ebb[ \blambda_{i}^{\mathrm{temp}} ] = \blambda_i^{*}$.
We can write the update
$\blambda_i^{(t)} \leftarrow \blambda_i^{*}$ recursively using the sample average
\begin{align*}
\blambda_i^{(t)} 
\leftarrow \frac{1}{t} \sum_{\tau = 1}^{t} \blambda_{i}^{\mathrm{temp},\tau}
& = \left(1 - \frac{1}{t}\right) \left( \frac{1}{t-1} \sum_{\tau = 1}^{t-1} \blambda_{i}^{\mathrm{temp},\tau} \right) + \frac{1}{t} \blambda_{i}^{\mathrm{temp},t} \\
& = \left(1 - \frac{1}{t}\right) \blambda_{i}^{(t-1)} + \frac{1}{t} \blambda_{i}^{\mathrm{temp},t} \ . 
\end{align*}
Define $\rho_t \defined \frac{1}{t}$.
In the running average, 
$\sum_{t = 1}^{\infty} \frac{1}{t} = \infty$
and $\sum_{t = 1}^{\infty} \left(\frac{1}{t}\right)^2 < \infty$,
and therefore $\sum_{t = 1}^{\infty} \rho_t = \infty$
and $\sum_{t = 1}^{\infty} \rho_t^2 < \infty$.
In the running average with $\rho_t \defined \frac{1}{t}$, each gradient sample is treated equally.
However, instead of incorporating fraction $\frac{1}{t}$ of $\blambda_{i}^{\mathrm{temp},t}$ into the
running average, we may erase a bit more from the ``past memory'' $\blambda_{i}^{(t-1)}$ to include a bit \emph{more}
of the recent gradient $\blambda_{i}^{\mathrm{temp},t}$. How much more is permissible?

Now define $\rho_t \defined t^{-\kappa}$.
For $\kappa = \frac{1}{2}$, the previous samples will be forgotten at a faster rate, and more of
$\blambda_{i}^{\mathrm{temp},t}$ will be included through
$\blambda_i^{(t)} \leftarrow (1 - t^{-1/2}) \blambda_{i}^{(t-1)} + t^{-1/2} \blambda_{i}^{\mathrm{temp},t}$.
However, at this rate past samples are forgotten too quickly, as both
$\sum_{t = 1}^{\infty} t^{-1/2} = \infty$ and $\sum_{t = 1}^{\infty} (t^{-1/2} )^2 = \infty$.
For any $\kappa' > 1$, both infinite sums will be finite,
e.g.~$\sum_{t = 1}^{\infty} t^{-\kappa'} < \infty$ and
$\sum_{t = 1}^{\infty} (t^{-\kappa'} )^2 < \infty$, and the running average will cling on to old memories, and has too little capacity
to incorporate recent gradient samples $\blambda_{i}^{\mathrm{temp},t}$.
Between forgetting too quickly or not at all, a setting of $\kappa \in (\frac{1}{2}, 1]$ in $\rho_t \defined t^{-\kappa}$ is therefore permissible.

%%%%%%%%%%%%%%%%%%%%%%%%%%%%%%%%%%%%%%%%%%%%%%
\subsection{Converging with fickle neighbours}

The running average in Sec.~\ref{sec:stochastic-grad} can boldly start at $\rho_t \defined t^{-\kappa} = 1$ for $t=1$,
and from this unit length step along the natural gradient,
accumulate gradient samples until convergence. However, it rests on the premise that neighbours
$\blambda_j$ for $j \neq i$ from the Markov blanket of $\x_i$ remain unchanged.

If this premise does not hold, much smaller steps $\rho_t \defined (t + \tau)^{-\kappa}$
with delay $\tau \ge 0$ are required.
This is indeed the case.
As Sec.~\ref{sec:bmf} shows,
a delay $\tau \ge 0$ that is sufficiently large for the stochastic gradient scheme to converge in practice
is not known \emph{a priori}.
By explicitly stating the shorthand definitions of $\widetilde{\bmeta}_i$ and $\widetilde{\bmeta}_{ki}$
in (\ref{eq:L-of-lambda-i}),
it is clear that the other $\blambda_j$ appear through multi-linear functions in
\begin{align*}
\Lcal(\blambda_i) & = \left(
\widetilde{\bmeta}_i \Big( \big\{ \Ebb_{j} [ \phi_j(\x_j) ] \big\}_{j \in \pa(i)} \Big) 
+ \sum_{k \in \ch(i)} \widetilde{\bmeta}_{ki} \Big( \Ebb_{k} [ \phi_k(\x_k) ], \,  \big\{ \Ebb_{j} [ \phi_j(\x_j) ] \big\}_{j \in \cp(i)} \Big) 
\right)^T \Ebb_i \left[ \phi_i(\x_i) \right] \\
& \quad\quad - \blambda_i^T \Ebb_i \left[ \phi_i(\x_i) \right] - \tilde{g}_i(\blambda_i) + \mathrm{const} \ .
\end{align*}
If we now consider the global gradient
$\nabla_{\bLambda} \Lcal(\bLambda) = [\nabla_{\blambda_1} \Lcal(\blambda_1), \ldots, \nabla_{\blambda_I} \Lcal(\blambda_I)]$,
it is clear from the above form (multi-linear in all variables)
that we can't set the gradient to zero and solve for all $\bLambda$ explicitly,
as was done in (\ref{eq:vb-update}). It is usually not even convex problem.

The gradient steps are along the global natural gradient. It is defined as
\[
\widetilde{\nabla}_{\bLambda} \Lcal(\bLambda) \defined G(\bLambda)^{-1} \nabla_{\bLambda} \Lcal(\bLambda) \ ,
\]
with $G(\bLambda)$ being the Fisher information matrix of $q$,
\[
G(\bLambda)
= \Ebb_q \left[ \Big(\nabla_{\bLambda} \log q(\X^\hidden | \bLambda) \Big) \Big( \nabla_{\bLambda} \log q(\X^\hidden | \bLambda) \Big)^T \right] \ .
\]
$G(\bLambda)$ is block-diagonal, as the covariance between $\phi_i(\x_i)$ and $\phi_j(\x_j)$ is zero for $i \neq j$, due to
the factorization of $q$.
Its inverse is therefore also block-diagonal,
and the natural gradient has the form
$\widetilde{\nabla}_{\bLambda} \Lcal(\bLambda) = [\widetilde{\nabla}_{\blambda_1} \Lcal(\blambda_1), \ldots, \widetilde{\nabla}_{\blambda_I} \Lcal(\blambda_I)]$.

%%%%%%%%%%%%%%%%%%%%%%%%%%%%%%
\section{Global batch samples}

An alternative to Alg.~\ref{alg:stochastic} is to take a batch data sample $\Dcal$ of $C_{\mathrm{global}}$ observed variables at the start of each iteration, and follow either each $\widetilde{\nabla}_{\blambda_i} \Lcal(\blambda_i)$ or $\widetilde{\nabla}_{\bLambda} \Lcal(\bLambda)$. This is outlined in Alg.~\ref{alg:stochastic-global}.

\begin{wrapfigure}[14]{L}[0pt]{0pt}
\noindent\begin{minipage}[t]{0.51\textwidth}
\vspace{-2pt}
\begin{algorithm}[H]
   \caption{Stochastic Variational Bayes}
   \label{alg:stochastic-global}
\begin{algorithmic}[1]
   \FOR{$t=1$ to $t_{\max}$ or convergence}
   \STATE $\rho_t = (t + \tau)^{-\kappa}$
   \STATE $\Dcal \leftarrow C_{\mathrm{global}}$ random nodes from $\X^\observed$
   \FOR{each hidden $\x_i$ : $i \in \pa(\Dcal)$}
   \STATE $\Dcal_i \leftarrow \{ \x_k \in \Dcal \cap \ch_i \}$
   \STATE $\blambda_i^{\mathrm{temp}} \leftarrow \widetilde{\bmeta}_i + \frac{N_i}{|\Dcal_i|} \sum_{k \in \Ccal}  \widetilde{\bmeta}_{ki}$
   \STATE \emph{option (a):} $\blambda_i \leftarrow (1 - \rho_t ) \blambda_i + \rho_t \blambda_i^{\mathrm{temp}}$
   \ENDFOR 
   \STATE \emph{// updates of $\pa_i$ etc.}
   \STATE \emph{option (b):} $\bLambda \leftarrow (1 - \rho_t ) \bLambda + \rho_t \bLambda^{\mathrm{temp}}$
   \ENDFOR
\end{algorithmic}
\end{algorithm}
\end{minipage}
\end{wrapfigure}

\begin{figure}[t]
%\vskip 0.2in
\begin{center}
\includegraphics[width=0.49\textwidth]{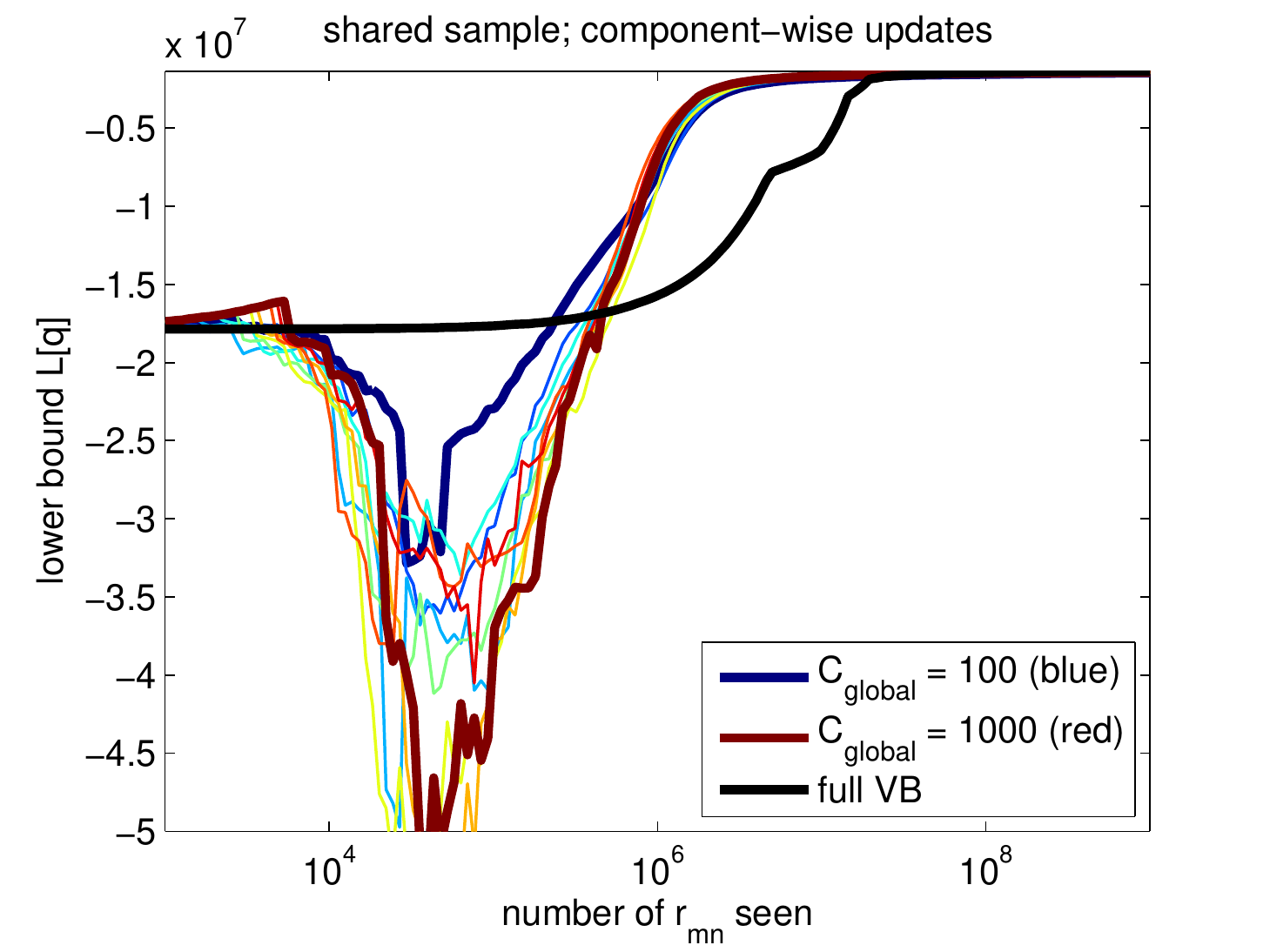}
\includegraphics[width=0.49\textwidth]{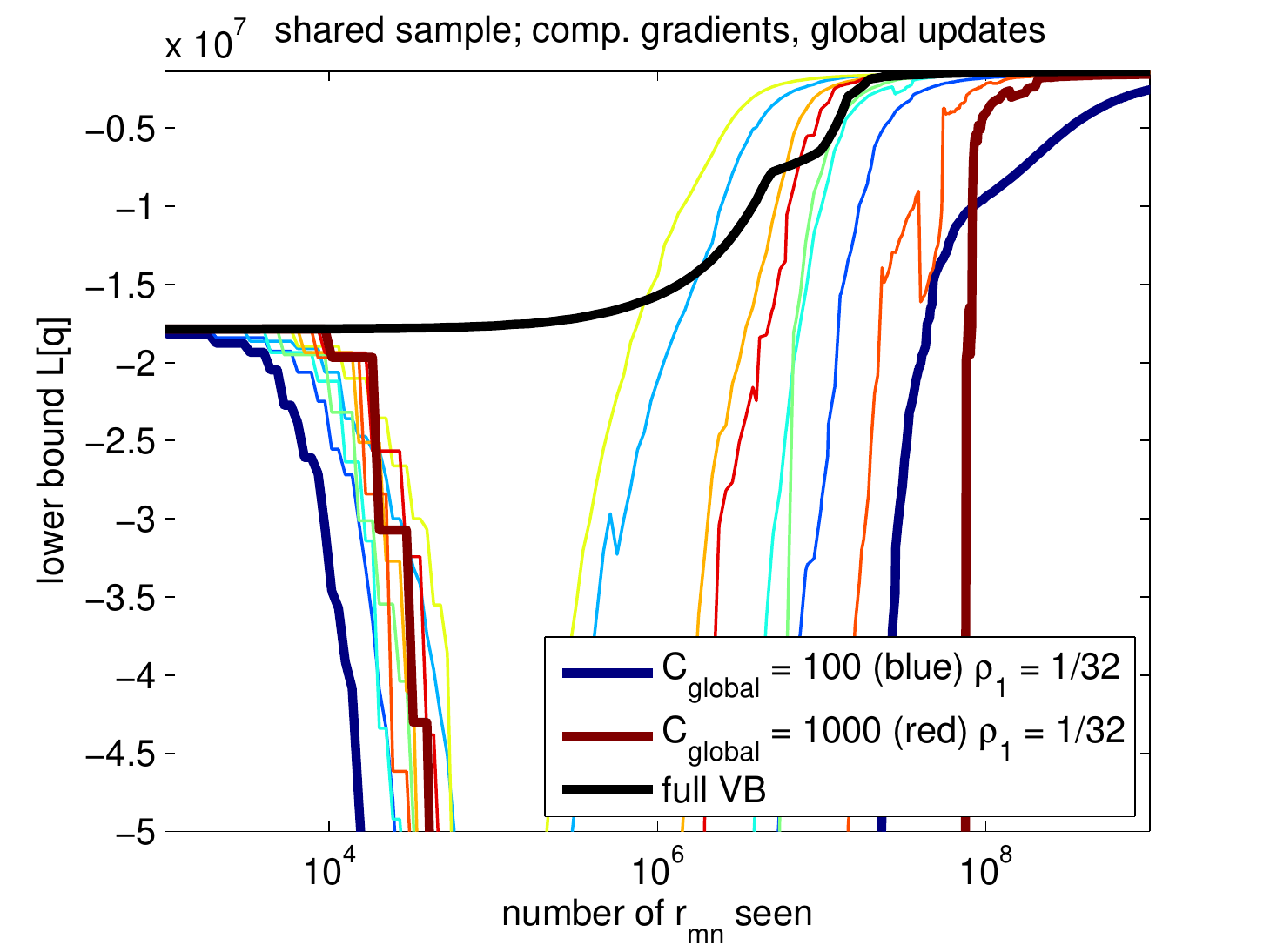} \\
\includegraphics[width=0.49\textwidth]{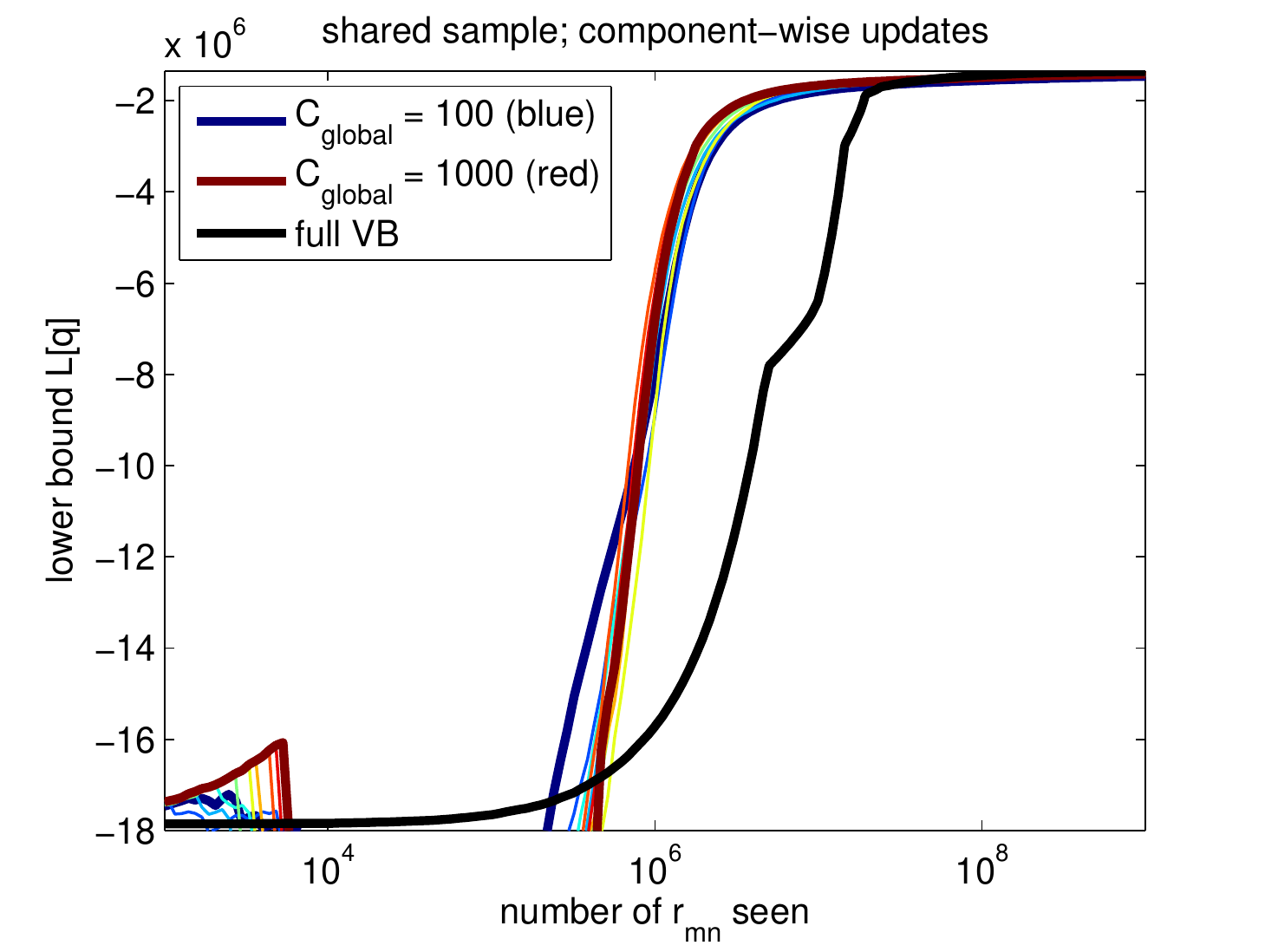}
\includegraphics[width=0.49\textwidth]{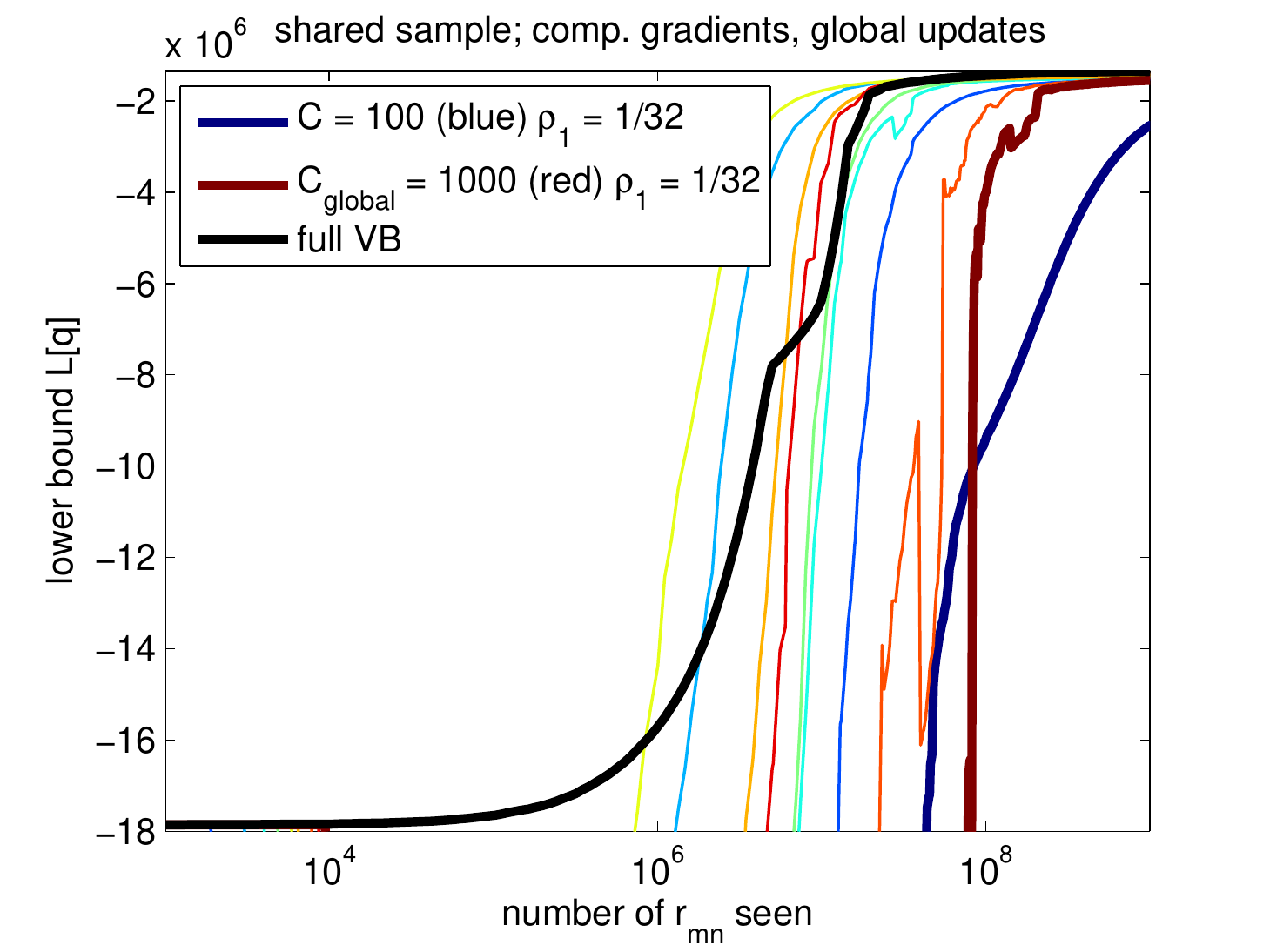} \\
\includegraphics[width=0.49\textwidth]{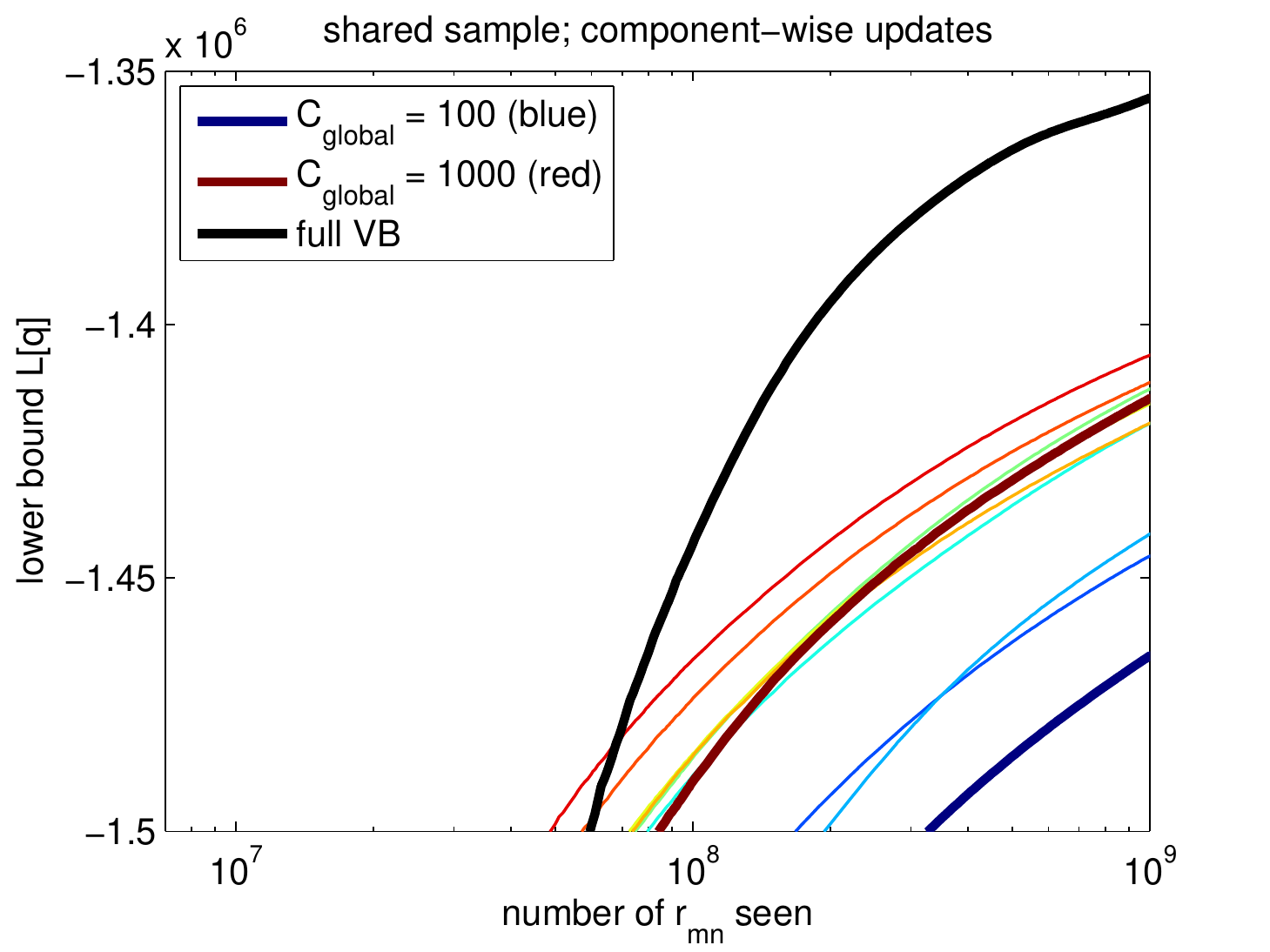}
\includegraphics[width=0.49\textwidth]{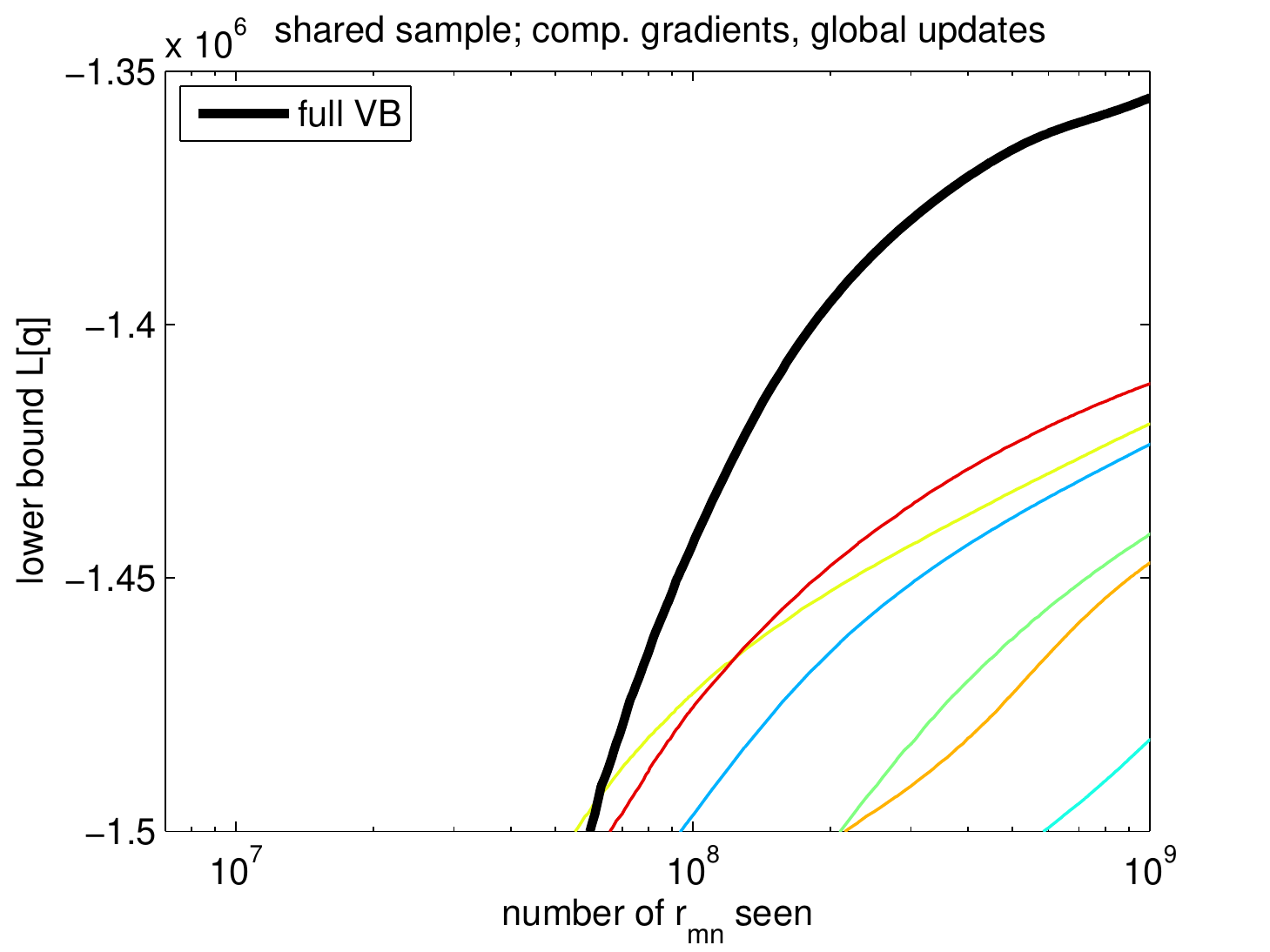}
\end{center}
\caption{Convergence of ${\cal L}[q]$ with $\rho_t = t^{-0.6}$. Alg.~\ref{alg:stochastic-global}'s option \emph{(a)} is shown in the left
column; option \emph{(b)} is shown in the right column. The x-axes are on a logarithmic scale.
}
\label{fig:convergenceGlobal}
\vspace{-10pt}
\end{figure} 
Fig.~\ref{fig:convergenceGlobal} considers batch sizes of $C_{\mathrm{global}} = 100$ to 1000, in intervals of 100.
The convergence in Fig.~\ref{fig:convergenceGlobal} is much slower than that of Fig.~\ref{fig:convergence}.
For the global update in option \emph{(b)} in Alg.~\ref{alg:stochastic-global}, the algorithm only converged on finite machine precision when $\rho_1 \le \frac{1}{32}$ was chosen (smaller for some  $C_{\mathrm{global}}$ settings), whereas for option \emph{(a)} at least the algorithm
converged from $\rho_1 = 1$ for all settings.

In Alg.~\ref{alg:stochastic-global}, $\pa(\Dcal)$ is the set of hidden variables that have a child node in $\Dcal$. Line 9 makes provision for updating variables (like hyper-parameters) that don't have observed children; this was not required for our example.

\end{document}